\pdfoutput=1

\documentclass[11pt]{article}

\usepackage{acl}

\usepackage{times}
\usepackage{latexsym}
\usepackage[T1]{fontenc}
\usepackage[utf8]{inputenc}
\usepackage{microtype}
\usepackage{inconsolata}
\usepackage{graphicx}
\usepackage{amssymb}
\usepackage{times}
\usepackage{latexsym}
\usepackage{diagbox}
\usepackage{multirow}
\usepackage{ulem}
\usepackage{pdfrender}
\usepackage{url}
\usepackage{enumitem}
\usepackage{cancel}
\usepackage{longtable}
\usepackage{times}
\usepackage{latexsym}
\usepackage{enumitem}
\usepackage[T1]{fontenc}
\usepackage[utf8]{inputenc}
\usepackage{microtype}
\usepackage{inconsolata}
\usepackage{times}
\usepackage{latexsym}
\usepackage{comment}
\usepackage{adjustbox}
\usepackage{booktabs}
\usepackage{subcaption}
\usepackage[english]{babel}
\usepackage[utf8]{inputenc}
\usepackage{amsmath}
\usepackage{graphicx}
\usepackage{epsfig}
\usepackage{authblk}
\usepackage{booktabs}   
\usepackage{multirow}   
\usepackage{array}      
\usepackage{caption}    

\title{PreP-OCR: A Complete Pipeline for\\ Document Image Restoration and Enhanced OCR Accuracy}

\author{
\bf  Shuhao Guan\textsuperscript{1},
  Moule Lin\textsuperscript{2},
  Cheng Xu\textsuperscript{1},
  Xinyi Liu\textsuperscript{1}\\
\bf  Jinman Zhao\textsuperscript{3},
  Jiexin Fan\textsuperscript{2},
  Qi Xu\textsuperscript{4},
  Derek Greene\textsuperscript{1}\\
  
  \textsuperscript{1}University College Dublin,
  \textsuperscript{2}Trinity College Dublin \\ 
  \textsuperscript{3}University of Toronto,
   \textsuperscript{4}Shanghai University \\
   \texttt{shuhao.guan@ucdconnect.ie}, \texttt{derek.greene@ucd.ie}
}

\begin{document}
\maketitle

\begin{abstract}
This paper introduces PreP-OCR, a two-stage pipeline that combines document image restoration with semantic-aware post-OCR correction to enhance both visual clarity and textual consistency, thereby improving text extraction from degraded historical documents.
First, we synthesize document-image pairs from plaintext, rendering them with diverse fonts and layouts and then applying a randomly ordered set of degradation operations. An image restoration model is trained on this synthetic data, using multi-directional patch extraction and fusion to process large images. Second, a ByT5 post-OCR model, fine-tuned on synthetic historical text pairs, addresses remaining OCR errors.
Detailed experiments on 13,831 pages of real historical documents in English, French, and Spanish show that the PreP-OCR pipeline reduces character error rates by 63.9-70.3\% compared to OCR on raw images. Our pipeline demonstrates the potential of integrating image restoration with linguistic error correction for digitizing historical archives. \url{https://github.com/NikoGuan/PreP-OCR}
\end{abstract}

\section{Introduction}
\label{introduction}

In the era of massive document digitization, ensuring accurate text extraction from degraded images has become increasingly important \cite{shen2021layoutparser}. Many historical documents, scanned books, and archival materials suffer from various forms of degradation -- such as blur, noise, ink bleeding, and other artifacts -- due to aging and suboptimal scanning conditions \cite{pardo2024advanced}. These degradations not only affect the visual quality of the images, but can also severely impact the resulting performance of Optical Character Recognition (OCR) systems, leading to high error rates in extracted text \cite{hegghammer2022ocr}.

To address these challenges, this paper introduces PreP-OCR, a novel synthetic-data-driven two-stage pipeline that first restores degraded images for OCR-based text extraction and then enhances the extracted text through post-processing.


To effectively train the image restoration model, we employ a comprehensive synthetic data generation method that simulates realistic document degradation. First, we render clean text images with diverse typography, then we apply degradation operations in a randomized order with stochastic parameters (see Section~\ref{sec:predata}), yielding a richly varied dataset, allowing models to learn a robust mapping between the original degraded inputs and their clean counterparts. Additionally, we propose a multi-directional patch extraction and fusion strategy to efficiently process larger images and further enhance overall image quality (see Section~\ref{sec:fusion}). Figure~\ref{fig:goodcase} shows examples of the process.

Following image restoration, in the next step of our proposed pipeline the restored images are fed into an OCR system. Although restoration significantly reduces structural ambiguities, it may not fully eliminate OCR errors. To correct any residual recognition mistakes, we incorporate a ByT5 post-OCR correction module that semantically recovers errors, even in cases where images are severely degraded and challenging to fully restore (see Section~\ref{sec:post}). Consequently, the restoration stage primarily resolves ambiguities in character shapes, yielding more legible images that are easier for OCR systems to recognize, while the post-correction stage mitigates systematic OCR errors through sequence-to-sequence translation.

In Sections~\ref{sec:realdataset} and \ref{sec:experiment1}, we describe the collection of numerous degraded historical book images.  These images were scanned using various OCR systems, and we then constructed evaluation datasets with their corresponding ground truth texts. In Sections~\ref{sec:experiment2}--\ref{sec:experiment3}, we use the data to assess text reconstruction quality in different patch regions and evaluate the effectiveness of our fusion strategy. Finally, in Sections~\ref{sec:experiment4}--\ref{sec:experiment5}, we test the PreP-OCR pipeline on English, French, and Spanish datasets.

\section{Related Work}
\label{related_work}
Extensive research has demonstrated that image pre-processing can significantly improve the performance of deep learning models \cite{vidal2012pre,salvi2021impact}. However, pre-processing within the context of OCR remains relatively underexplored, with existing methods primarily focusing on contrast enhancement and color adjustment \cite{gupta2007ocr,harraj2015ocr,bui2017selecting}.

Recent studies in image deblurring have introduced more advanced restoration techniques that could also benefit OCR. Early image restoration methods were primarily based on CNNs \cite{dong2015compression,dong2015image,zhang2017beyond,cho2021rethinking}. Subsequent research introduced more elaborate architectures, such as residual blocks \cite{kim2016accurate,zhang2021plug}, generative adversarial networks (GANs) \cite{pathak2016context,gulrajani2017improved,wang2018esrgan,kupyn2019deblurgan}, and attention mechanisms \cite{zhang2018image,yu2018generative}. Transformers \cite{vaswani2017attention}, which model long-range dependencies, have advanced NLP and computer vision and are now widely used in image restoration \cite{chen2021pre,liang2021swinir,zamir2022restormer}.

Diffusion models have emerged as a powerful alternative for generative image tasks, optimizing a parameterized Markov chain to approximate the target distribution more accurately than many other generative frameworks. Examples in restoration include DiffIR \cite{xia2023diffir} and ResShift \cite{yue2024resshift}, both of which are diffusion-based approaches. Several studies have also used diffusion models together with textual information to recover the appearance of ancient stele inscriptions \cite{10.1145/3664647.3680587,Yang_Peng_Shi_Zhang_Liu_Jin_2025}. In our work, we harness image-restoration models to pre-process degraded images prior to applying OCR.

The post-OCR task aims to correct errors in OCR outputs, with early methods relying on dictionary lookups or spelling checkers \cite{furrer2011reducing,bassil2012ocr,estrella2014ocr,kettunen2016keep}. More recent approaches treat post-OCR correction as a sequence-to-sequence task, leveraging neural machine translation (NMT) models, such as BERT \cite{devlin2019bertpretrainingdeepbidirectional}, BART \cite{lewis2019bart} and T5 \cite{raffel2020exploring} \cite{amrhein2018supervised,nguyen2020neural,soper2021bart,maheshwari2022benchmark}. Several comparative studies have shown that byte-level models, such as ByT5 \cite{xue2022byt5}, often achieve the best performance for post-OCR tasks \cite{maheshwari2022benchmark,lofgren2024post,guan2024effective,guan2024synthetically}.

Both image restoration and post-OCR correction require paired training data, and the availability of abundant, high-quality data is critical for success \cite{rijhwani-etal-2020-ocr,mazumder2024dataperf,bi2025prismselfpruningintrinsicselection}. Consequently, researchers have explored a variety of strategies for generating synthetic data as a form of data augmentation \cite{hamdi2023depth,shorten2019survey}. For image deblurring and text-recognition, common techniques involve injecting noise into clean images to mimic real-world degradation \cite{yuan2007image,krishna-etal-2018-upcycle,rim2022realistic,li2023image,hamdi2023depth}, or using methods such as StableDiffusion \cite{rombach2022high} to create paired image edits \cite{brooks2023instructpix2pix}. In the post-OCR domain, synthetic training pairs are often produced by inserting controlled errors into clean text \cite{dhondt-etal-2017-generating,grundkiewicz2019neural,ignat2022ocr,jasonarson2023generating,guan2024advancing,guan2024effective}.

\section{Problem Formulation}
\label{sec:formulation}
Our task addresses two sequential objectives: (1) restoring degraded images to enhance legibility, and (2) recovering accurate textual content from these images. We formalize these goals as follows.

\vskip 0.4em
\noindent{\textbf{Image restoration objective.}}
Let \( I_d, I \in \mathbb{R}^{H \times W} \) denote the degraded input and its sharp ground-truth image, respectively. A restoration model \( \mathcal{R} \) aims to produce a restored image \( \hat{I} = \mathcal{R}(I_d) \), where the objective is to maximize the Peak-Signal-to-Noise Ratio (PSNR) \cite{hore2010image} between $\hat{I}$ and $I$, such that:
\begin{equation*}
\mathcal{R}^* = \arg\max_{\mathcal{R}} \ \text{PSNR}(\mathcal{R}(I_d), I),
\end{equation*}
%
\vskip -0.2em
\noindent{\textbf{Text recovery objective.}}
Let \( T \) represent the ground-truth text sequence of image \( I_d \). The restored image \( \hat{I} \) is first processed by an OCR model \( \mathcal{O} \), yielding predicted text \( T' = \mathcal{O}(\hat{I}) \). This predicted text \( T' \) is then refined by a post-processing module \( \mathcal{P} \), resulting in \( \hat{T} = \mathcal{P}(T') \). The objective here is to minimize the Character Error Rate (CER) between \( \hat{T} \) and \( T \):
\begin{equation*}
\mathcal{P}^* = \arg\min_{\mathcal{P}} \ \text{CER}(\mathcal{P}(\mathcal{O}(\hat{I})), T),
\end{equation*}
These dual objectives are addressed in our two-stage pipeline. First, the restoration model $\mathcal{R}$ is optimized using synthetic paired data to restore the book images, directly enhancing character legibility (see Section~\ref{sec:predata}). Second, the post-processor $\mathcal{P}$ is trained on synthetic training pairs simulating OCR errors to correct residual recognition mistakes (see Section~\ref{sec:post}). The image restoration stage reduces structural ambiguities in character shapes, while the text correction stage addresses systematic OCR errors through sequence-to-sequence translation. This cascaded approach ensures both pixel-level fidelity in $\hat{I}$ and semantic-level accuracy in the final text output $\hat{T}$.

\section{PreP Pipeline}

\subsection{Real Evaluation Data Collection}
\label{sec:realdataset}

\begin{figure}[htbp]
    \centering
    \includegraphics[width=0.48\textwidth]{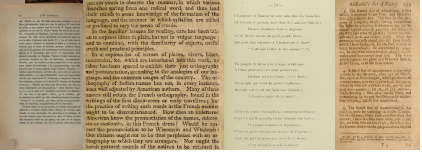} 
    \caption{Example images of digitized pages from historical books, which are often affected by degraded text, aging pages, and low capture resolution.} 
    \label{fig:book} 
\end{figure}

To evaluate the performance of a model trained solely on synthetic data when applied to real-world data, we constructed a new corpus as follows. We curated a collection of 30 English books (9,606 pages), 5 Spanish books (2,404 pages), and 5 French books (1,821 pages) from the 15th to 19th centuries. Ground truth (GT) texts were sourced from clean digital books available on Project Gutenberg\footnote{\url{https://www.gutenberg.org}}, while a set of corresponding scanned PDF files containing degraded text images was obtained from Open Library\footnote{\url{https://openlibrary.org}}. We intentionally selected older books exhibiting visible damage, as shown in the images in Figure~\ref{fig:book}. Text alignment between the OCR outputs and GT was performed using the RETAS framework \cite{yalniz2011fast}, which employs dynamic programming for robust sequence matching.

For subsequent experiments, we pre-process the images through denoising before employing OCR. Comparative CER analysis will be conducted across three pipelines: raw images (direct OCR on original scanned pages), Pre-process (OCR after image restoration), and our proposed approach PreP-process (image restoration combined with OCR and post-correction).

\subsection{Synthetic Data for Restoration}
\label{sec:predata}

In image-to-image restoration tasks, paired data consisting of a degraded input and its corresponding clean reference is crucial for effective training. However, obtaining such paired data from real-world documents is extremely challenging because authentic clean images and their degraded counterparts are rarely available. To overcome this limitation, we employ a synthetic data generation approach that enables us to simulate realistic degradation from scratch.

Our synthetic data pipeline begins by generating a clean base image from textual content. To maximize OCR accuracy, we ignore color information and work with grayscale images. First, we collect various fonts for different languages and render multi-line text with a range of stylistic variations, including random indentation, character shifts, rotation, and bending. Additionally, the text is randomly tilted, and both line and character spacing are varied to mimic the natural irregularities found in printed documents. The generated base image serves as the clean ground truth.

To simulate real-world degradations, next we apply a series of controlled noise and distortion operations. Specifically, the pipeline adds random noise, performs resolution reduction, applies Gaussian blurring, and overlays additional artifacts such as random black or white patches of varying sizes, white or black lines (simulating scratches or folds), background textures, and stain overlays. The process also includes random morphological operations (dilation and erosion) to further simulate text imperfections. It is worth noting that these operations are applied in random order, producing diverse results depending on the sequence.

Since noise levels can vary in real-world digitized documents, we predefine four noise levels (level-1 to level-4). Higher levels introduce a wider range of noise parameters, potentially resulting in more degraded images. Additionally, 10\% of the noisy images are binarized using Otsu's algorithm \cite{yousefi2011image}. We also stitch together images with different noise levels and fonts, as in real data, different regions on a given page can exhibit varying degrees of degradation and typographic styles. 

\begin{figure}[!t]
    \centering
    \includegraphics[width=0.48\textwidth]{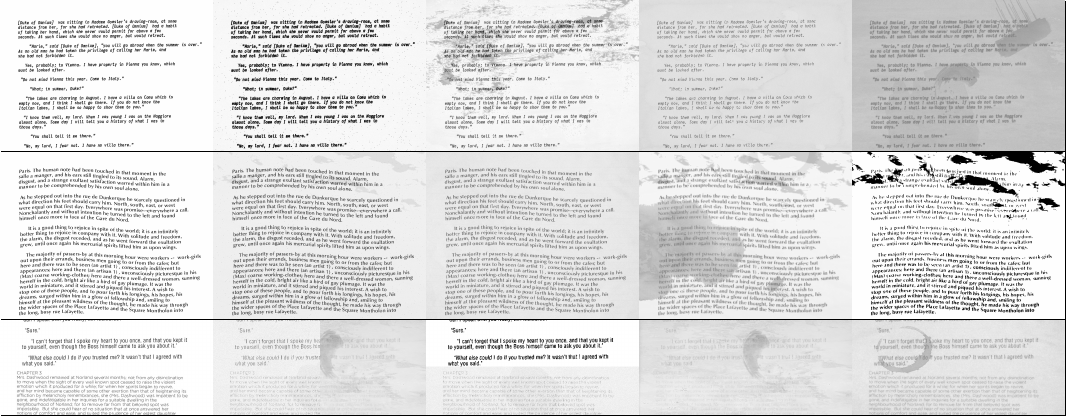} 
    \caption{Example of three sets of synthetic image data. The leftmost image is the base image, while the image to its right is the corresponding degraded image.} 
    \label{fig:pdf} 
\end{figure}

The detailed parameters for generating the base image and simulating noise levels are provided in Appendix \ref{sec:appendix_noise}. Example images generated using this process are shown in Figure~\ref{fig:pdf}.

By pairing each original base image with its synthetically degraded versions, we create a large and diverse dataset. This synthetic data facilitates the robust training of our restoration model, allowing it to learn the complex mapping from degraded to clean images. As demonstrated later in Section~\ref{results}, this can ultimately improve generalization performance in real-world document restoration tasks.

\subsection{Patch Extraction and Fusion}
\label{sec:fusion}

When processing large images, we first partition them into multiple regions. To address stochastic noise and local inconsistencies, we adopt a multi-directional patch extraction strategy. Specifically, for each degraded image, we scan it four times: top-left to bottom-right, top-right to bottom-left, bottom-left to top-right, and bottom-right to top-left. Since image dimensions may not align perfectly with the patch stride, we pad only the edge opposite the scanning direction to ensure a fully integer-aligned pass over the entire image. 

In each pass, 256×256 patches are extracted at a stride of 128 pixels. Scanning from different directions yields slightly different patches, meaning even the same region in the original image may appear with different neighboring contexts in a patch—leading to varied predictions. After restoring each patch, we discard the outer 64-pixel border and retain only the central 128×128 region, minimizing boundary artifacts. An example of the multi-direction patch extraction process is provided in Appendix \ref{sec:appendix_patch}.

\begin{figure}[!t]
    \centering
    \includegraphics[width=0.48\textwidth]{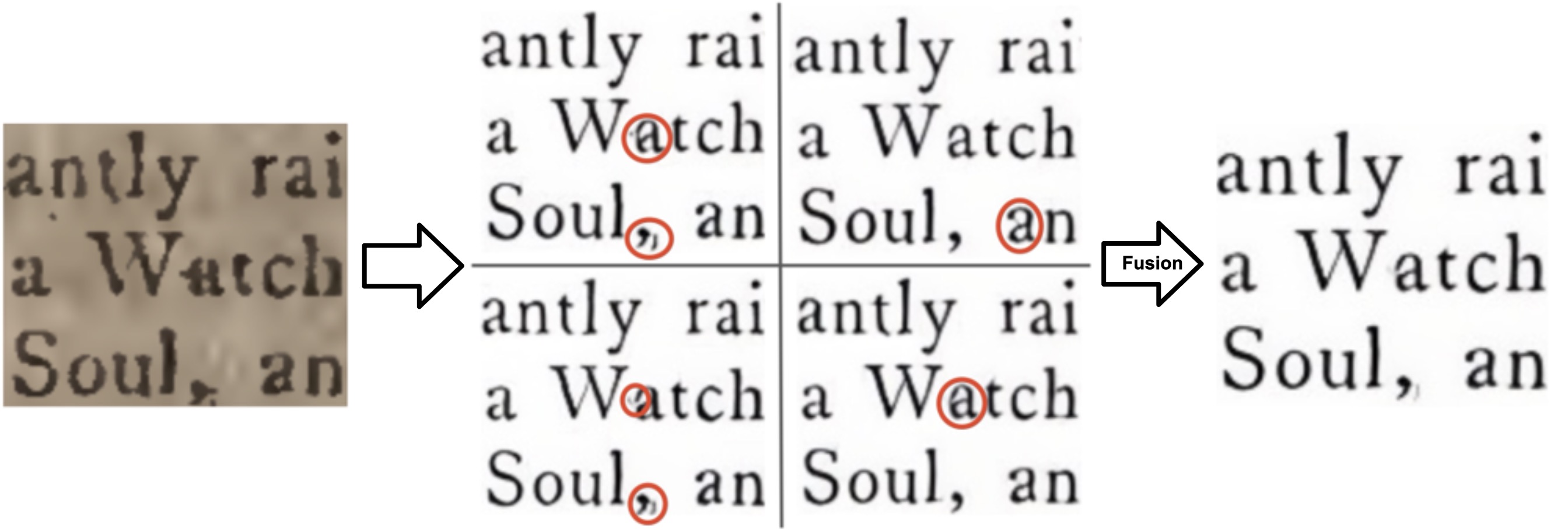} 
    \caption{The left panel shows a real degraded patch. The four sub-panels in the center depict restored outputs under different scanning directions, where the red circles highlight localized artifacts or noise. On the right is the final fused result, in which these artifacts are effectively suppressed.} 
    \label{fig:fusion} 
\end{figure}

Each pixel in the final restored image is fused by aggregating four independent predictions from the four scanning directions. Specifically, for each scanning direction \( k \in \{1,2,3,4\} \), the restoration model \( \mathcal{R} \) generates an intermediate restored image \( \hat{I}_k \). To merge these predictions and reduce artifacts, we perform a pixel-wise median operation across the four resulting images. Formally, the final restored image \( \hat{I} \) is computed as
\[
\hat{I}[r,c,\chi] = \mathrm{median}\bigl(\hat{I}_k[r,c,\chi] \mid k \in \{1,2,3,4\}\bigr)
\]
where \( \chi  \)  is the grayscale intensity, $r$ and $c$ are the row and column indices. This median operation consolidates the consistent pixel values across different scanning paths, improving the stability and quality of the final restored image. As shown in Figure \ref{fig:fusion}, the median fusion suppresses outlier predictions caused by artifacts and stochastic noise.

Our multi-directional scanning strategy aggregates predictions from overlapping patches processed through varied spatial contexts, analogous to multi-view consensus mechanisms in image processing. This approach enhances OCR outputs, as demonstrated later in Section \ref{sec:experiment3}.

\subsection{Post-OCR Correction}
\label{sec:post}

Building on the image restoration pipeline described in Sections~\ref{sec:predata}--\ref{sec:fusion}, our pipeline incorporates a post-processor to address residual OCR errors. While the pre-processing stage enhances text legibility, characteristic OCR mistakes persist due to (1) morphological ambiguities in restored characters, and (2) linguistic context gaps in OCR engines. To mitigate these, we implement an error correction module based on \citet{guan2024effective}'s synthetic data approach, adapted to our pre-processing outputs.

We first extract the OCR error distribution from a small post-OCR dataset -- the ICDAR 2017 post-OCR data \cite{chiron2017icdar2017}. Then, we inject  errors into clean text to generate a large-scale synthetic training pair  \( (T, T') \), the ByT5-base model \cite{xue2022byt5} $\mathcal{P}$ is then trained to map \( T' \) to \( T \), leveraging byte-level tokenization to handle rare characters from historical documents.

Specifically, we simulate OCR errors by replacing characters in the clean text \( T \) according to error distributions derived from the ICDAR. For example, the letter “m” might have an error set such as \{"n": 0.001, "rn": 0.002, …\}, where each error candidate is assigned an occurrence probability. These error sets may include various symbols, spaces, multi-character sequences, and the placeholder “@”. We uniformly adjust the overall error rate so that, as the error rate increases, characters are more likely to be replaced by an erroneous element, leading to a higher CER. After the replacement process, any placeholders are removed from the text. This procedure can simulate recognition, insertion, deletion, and segmentation errors.

This design complements our image restoration stage: while Section~\ref{sec:fusion}'s fusion reduces local artifacts, the post-processor resolves systemic OCR errors through learned linguistic patterns. The combined PreP-OCR pipeline thus addresses both visual ambiguities (via $\mathcal{R}$) and semantic inconsistencies (via $\mathcal{P}$), as we observe later in Section~\ref{sec:experiment4}.




\section{Experiments}
\label{results}

\subsection{Exp.~1: OCR Performance}
\label{sec:experiment1}

In our first experiment, we evaluate OCR performance on the real book dataset described in Section \ref{sec:realdataset}. While Tesseract has been the most widely used OCR engine  \cite{smith2007overview}, recent advances in Transformer-based models have led to the emergence of general-purpose large language models (LLMs) with strong visual capabilities \cite{xu2024benchmark,bi2025llavasteeringvisualinstruction,bi2025cotkineticstheoreticalmodelingassessing,yu2025table,chen2025c}, as well as specialized LLMs for OCR.

For baseline evaluation, we employ three OCR systems: Tesseract-5.5.0 \cite{smith2007overview}; GOT \cite{wei2024general}, a LLM designed for OCR tasks; and GPT-4o-2024-08-06 \cite{openai2024gpt4ocard}. Details are provided in Appendix \ref{sec:appendix_gpt}. We used the RETAS framework \cite{yalniz2011fast} to align the OCR outputs with the GT text. After alignment, we computed the Character Error Rate (CER) and Word Error Rate (WER) to assess each system's accuracy. Since text extracted from PDFs often contains extraneous content that is not part of the main body, any text segments that do not have a corresponding match in the GT were discarded and excluded from the CER calculation.

\begin{table}[htbp]
\centering
\resizebox{0.48\textwidth}{!}{%

\begin{tabular}{l|cc|cc|cc}
\toprule
\multirow{2}{*}{Model} & \multicolumn{2}{c}{English} & \multicolumn{2}{c}{French} & \multicolumn{2}{c}{Spanish} \\
\cmidrule(lr){2-3} \cmidrule(lr){4-5} \cmidrule(lr){6-7}
 & CER & WER  & CER  & WER  & CER & WER \\
\midrule
Tesseract & 5.91 (5.91) & 26.70 (26.70) & 5.16 (5.11)& 27.21 (26.97)& 7.12 (7.12)& 27.13 (27.13) \\
GOT       & 11.18 (6.95)& 35.12 (20.29) & 6.32 (5.15)& 28.53 (25.43)       &    12.84 (6.29)        &      46.10 (24.32)         \\
GPT-4o    & 6.51 (\textbf{2.34}) & 9.37 (\textbf{3.43})   & 3.23 (\textbf{1.93})      & 4.98 (\textbf{4.68})         &     3.43 (\textbf{1.84})       &  5.42(\textbf{2.00}) \\
\bottomrule
\end{tabular}%
}
\caption{Character Error Rate (CER) and Word Error Rate (WER) across Languages and Models, the values in parentheses are the results obtained after removing abnormal pages with a CER greater than 25\%. Boldface indicates the best performance in each metric for each language.}
\label{tab:ocr_results}
\end{table}

Table \ref{tab:ocr_results} shows the final results. We observe that the LLM-based OCR systems are less stable than Tesseract, often producing outliers characterized by incomplete page outputs or extraneous content. However, after removing these outlier pages (i.e., CER >25\%), GPT-4o performs very well. In contrast, GOT remains unstable and does not exhibit outstanding performance even after outlier removal. Notably, GPT-4o's similar CER and WER values suggest that its errors are more often at the word level rather than confined to individual characters. Further analysis of the CER distribution for English and additional details are provided in Section \ref{sec:experiment4}.

\subsection{Exp.~2: Patch Restoration Assessment} 
\label{sec:experiment2}

In this experiment, we train and evaluate six image-to-image models on synthetic data generated according to Section~\ref{sec:predata}: ResShift \cite{yue2024resshift}, DeblurGAN-v2 \cite{kupyn2019deblurgan}, MIMO-UNet+ \cite{cho2021rethinking}, DiffIR \cite{xia2023diffir}, Restormer \cite{zamir2022restormer}, and IP2P \cite{brooks2023instructpix2pix}. We created a total of 100{,}000 image pairs, of which 90{,}000 are used for training, 5{,}000 for validation, and 5{,}000 for testing. Each model is trained on randomly cropped $256 \times 256$ patches from the training set, training parameters are in Appendix \ref{sec:appendix_imagepara}. For testing, we extract two fixed $256 \times 256$ patches from each test image to ensure a uniform and controlled comparison across models. Note that this experiment assesses only the patch-wise performance.

Our main evaluation on real data focuses on OCR outputs, discussed later in Section~\ref{sec:experiment3}. However, to directly assess how well these models reconstruct text regions and how border removal impacts performance, we use the synthetic test set and compute the Aggregated Masked PSNR (AMP). Specifically, we apply Otsu’s thresholding to both the ground-truth and the predicted patches to identify black text pixels, and then take the union of the two resulting masks to obtain $\mathcal{M}_U$. For each $(x,y)\in \mathcal{M}_U$,
\[
E(x,y) \;=\; \bigl(I(x,y) \;-\; \hat{I}(x,y)\bigr)^2.
\]
If $E(x,y)=0,$ we assign 100\,dB; otherwise,
\[
\text{PSNR}(x,y) \;=\;
10 \,\log_{10}\!\Bigl(\tfrac{255^2}{E(x,y)}\Bigr).
\]
This masking step excludes large uniform background regions so that the PSNR focuses on text fidelity.


We accumulate $\text{PSNR}(x,y)$ for every pixel $(x,y) \in \mathcal{M}_U$ across all test images, normalize by the number of times $(x,y)$ lies in $\mathcal{M}_U$. This yields an average map $\overline{\text{PSNR}}(x,y)$, where each pixel's value reflects its average PSNR across all relevant test patches' text region. If $\text{PSNR}_i(x,y)$ denotes the local PSNR for pixel $(x,y)$ in the $i$-th image, and $n(x,y)$ is the count of images where $(x,y)\in \mathcal{M}_U$:
\[
\overline{\text{PSNR}}(x,y) \;=\;
\frac{1}{n(x,y)}\,
\sum_{i=1}^{n(x,y)} \text{PSNR}_{i}(x,y).
\]
Finally, we compute AMP by taking the average of all pixel values in the $\overline{\text{PSNR}}(x,y)$:

\[
\text{AMP}
\;=\;
\frac{1}{|\Omega|}
\sum_{(x,y)\in \Omega}
\overline{\text{PSNR}}(x,y),
\]
where $\Omega$ is the set of all pixels in $\overline{\text{PSNR}}(x,y)$.

\begin{figure}[htbp]
    \centering
    \includegraphics[width=0.48\textwidth]{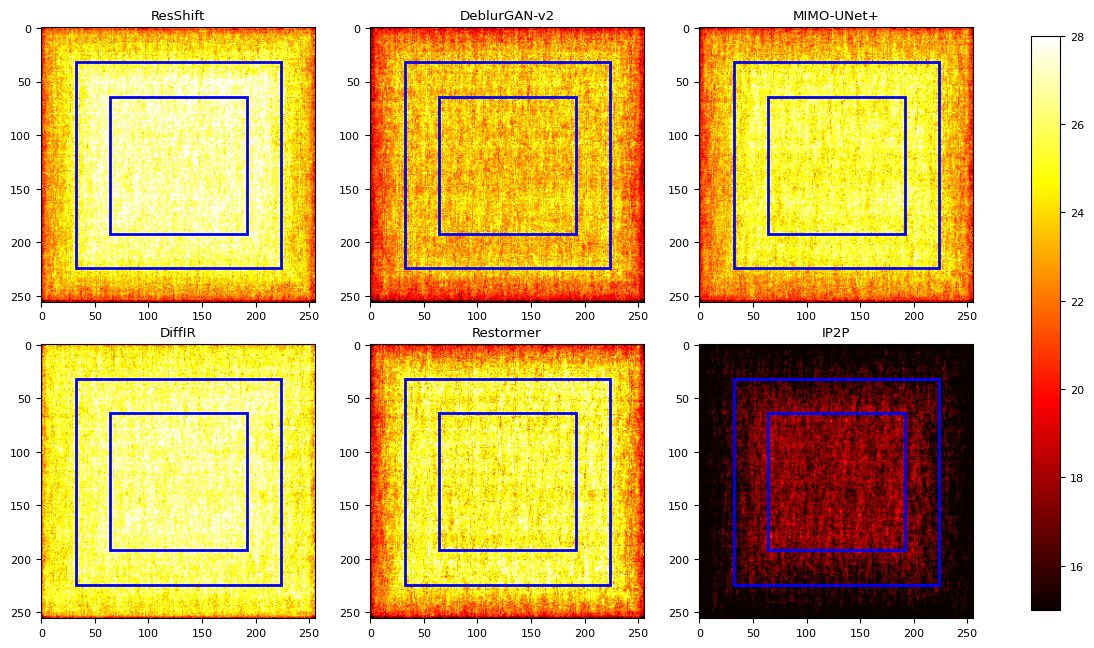}
    \caption{Visualization of $\overline{\text{PSNR}}$ for selected methods. The blue boxes highlight different regions within the images. Central regions tend to exhibit higher $\overline{\text{PSNR}}$.}
    \label{fig:amp}
\end{figure}

Table~\ref{tab:psnr_comparison} reports the AMP results and indicates that DiffIR achieves the highest AMP on full images (25.64\,dB), while ResShift performs well in the central subregions (26.58\,dB, 26.82\,dB). IP2P consistently underperforms. Figure~\ref{fig:amp} visualizes the $\overline{\text{PSNR}}$. The results indicate that the central regions generally achieve higher $\overline{\text{PSNR}}$ values compared to the border areas.

\begin{table}[!h]
\centering
\resizebox{0.48\textwidth}{!}{%
\small
\begin{tabular}{l | r r r}
\toprule
\multirow{2}{*}{Method} & \multicolumn{3}{c}{AMP $\uparrow$ (dB)} \\
\cmidrule(l){2-4}
 & Full Patch & Central-192 & Central-128 \\
\midrule
ResShift   &  25.18  & \textbf{26.58} & \textbf{26.82}   \\
DeblurGAN-v2   & 22.81 & 23.56  & 23.56  \\
MIMO-UNet+     & 24.08 & 25.26 & 25.40 \\
DiffIR    & \textbf{25.64} & 26.29 & 26.50 \\
Restormer      &24.13  & 25.29 & 25.18 \\
IP2P           & 17.14 & 17.29 & 17.35 \\
\bottomrule
\end{tabular}
}
\caption{AMP results for each restoration method, evaluated on the full 256$\times$256 patch and two central subregions (192$\times$192, 128$\times$128). Boldface highlights the best performance. Underlining indicates the best performance in each row.}
\label{tab:psnr_comparison}
\end{table}

\subsection{Exp.~3: Full-Page Restoration} 
\label{sec:experiment3}
Building on the synthetic-data evaluations in Experiment~2, we now investigate how reconstructed real historical images affect OCR performance. We also examine how Multi-Directional Patch Extraction combined with different fusion methods influences performance. Here, Tesseract is chosen for its stability; on the raw book images, it achieves a baseline CER of 5.91\%.

We resize each real degraded image $I_{d}$ to a width of 1216 pixels for consistency. Each model is tested under several configurations: Single-directional patch extraction (with 0, 32, or 64 pixels removed from each border) and multi-directional patch extraction using either median or mean fusion, again with 0, 32, or 64 border pixels removed. Table~\ref{tab:cer_comparison} shows the resulting CER for each configuration.

\begin{table*}[htbp]
\centering
\resizebox{\textwidth}{!}{%
\begin{tabular}{l|rrr|rrr|rrr}
\toprule
\multirow{2}{*}{Model} & \multicolumn{9}{c}{Configuration} \\
\cmidrule(lr){2-10}
 & Single-0 & Single-32 & Single-64 & Multi-Median-0 & Multi-Median-32 & Multi-Median-64 & Multi-Mean-0 & Multi-Mean-32 & Multi-Mean-64 \\
\midrule
ResShift
& 4.43 & \textbf{3.20} & 3.17
& 4.10 & \textbf{2.95} & \textbf{\underline{2.81}}
& 4.25 & \textbf{2.93} & 2.99 \\
DeblurGAN-v2
& 5.82 & 4.75 & 4.63
& 5.12 & 4.52 & \underline{4.48}
& 5.34 & 4.78 & 4.65 \\
MIMO-UNet+
& 4.65 & 3.89 & 3.70
& 4.22 & 3.68 & \underline{3.65}
& 4.41 & 3.82 & 3.77 \\
DiffIR
& \textbf{3.77} & 3.22 & \textbf{3.12}
& \textbf{3.63} & 3.10 & 2.94
& \textbf{3.52} & 3.23 & \textbf{\underline{2.91}} \\
Restormer
& 4.78 & 3.95 & 3.82
& 4.35 & 3.72 & 3.68
& 4.58 & 3.88 & \underline{3.60} \\
IP2P
& 54.35 & 59.42 & 49.28
& 46.01 & \underline{39.25} & 48.03
& 47.02 & 46.48 & 46.32 \\
\bottomrule
\end{tabular}
}
\caption{Character Error Rate (CER\%) across models and configurations. 
``Single-X'' indicates a single-directional patch extraction with X pixels removed from each border; 
``Multi-Median-X'' and "Multi-Mean-X'' indicate multi-directional fusion (median or mean, respectively). 
Boldface highlights the best performance in each column. Underlining indicates the best performance in each row.}
\label{tab:cer_comparison}
\end{table*}

From the results in Table~\ref{tab:cer_comparison}, we observe that median fusion generally outperforms mean fusion, while fusing multiple patches yields lower CER than using a single patch. Removing border pixels significantly improves performance, with 32-pixel removal already yielding a large gain and 64-pixel removal providing a modest further improvement. Under the Multi-Median-64 setting, ResShift achieves the best results, reducing the average CER by 52.45\% across 30 English books.

For the ResShift model, although truncating 64 pixels from each border of a 1024$\times$1024 image requires processing 64 patches in single-direction (11.3 seconds total) and 256 patches in multi-direction (45 seconds) on an RTX 4090, compared to 36 and 144 patches (6.36 and 25.46 seconds) for a 32-pixel truncation, the accuracy gain with Multi-Median-64 justifies the increased inference time. Consequently, we adopt Multi-Median-64 for our remaining experiments.

\subsection{Exp.~4: PreP-OCR Pipeline}
\label{sec:experiment4}

We now evaluate the complete PreP-OCR pipeline (image pre-processing, OCR, and post-processing) on real English book images. We investigate each step (i.e., pre-processing alone, and pre-processing combined with post-OCR correction) using the three OCR systems introduced in Section~\ref{sec:experiment1}.

\begin{figure}[!t]
    \centering
    \includegraphics[width=0.48\textwidth]{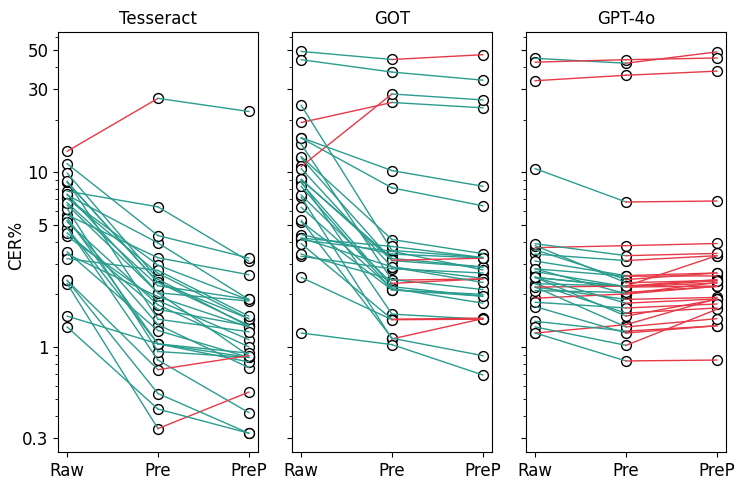} 
    \caption{CER values for each book in the real dataset under different processing pipelines for 3 OCR systems. The green line indicates a decrease in CER, while the red line indicates an increase.} 
    \label{fig:3ocr} 
\end{figure}

We selected 50 nineteenth-century British and Irish novels from Project Gutenberg, comprising 5,714,139 words. From these texts, we generated 894,271 synthetic training pairs (each up to 512 characters) to train the ByT5 post-correction model (see Appendix~\ref{sec:appendix_Post} for training details). The results are summarized in Table~\ref{tab:pipeline}, and Figure~\ref{fig:3ocr} visualizes the Character Error Rate (CER) across books for each pipeline configuration.

\begin{table}[!t]
  \centering
\resizebox{0.45\textwidth}{!}{%
  \begin{tabular}{l|rrr}
    \toprule
    \multirow{2}{*}{OCR Model} & \multicolumn{3}{c}{Pipeline} \\
    \cmidrule(l){2-4}
    & Raw & Pre & PreP \\
    \midrule

    Tesseract   & 5.91 (5.87) & 2.81 (\textbf{1.99}) & 2.00 (\textbf{\underline{1.30}}) \\
    GOT         & 11.18 (6.95) & 7.11 (3.00) & 6.65 (\underline{2.65}) \\
    GPT-4o      & 6.51 (\textbf{2.34}) & 6.06 (\underline{2.20}) & 6.57 (2.40) \\
    \bottomrule
  \end{tabular}%
  }
    \caption{CER of Tesseract, GOT, and GPT-4o under three pipelines: Raw (original images), Pre (ResShift pre-processing), and PreP (ResShift pre-processing + post-correction). Parentheses show CER after excluding outliers (i.e., pages where CER > 25\%). Boldface highlights the best performance in each column. Underlining indicates the best performance in each row.}
  \label{tab:pipeline}
\end{table}

\begin{figure*}[!t]
    \centering
    \includegraphics[width=\textwidth]{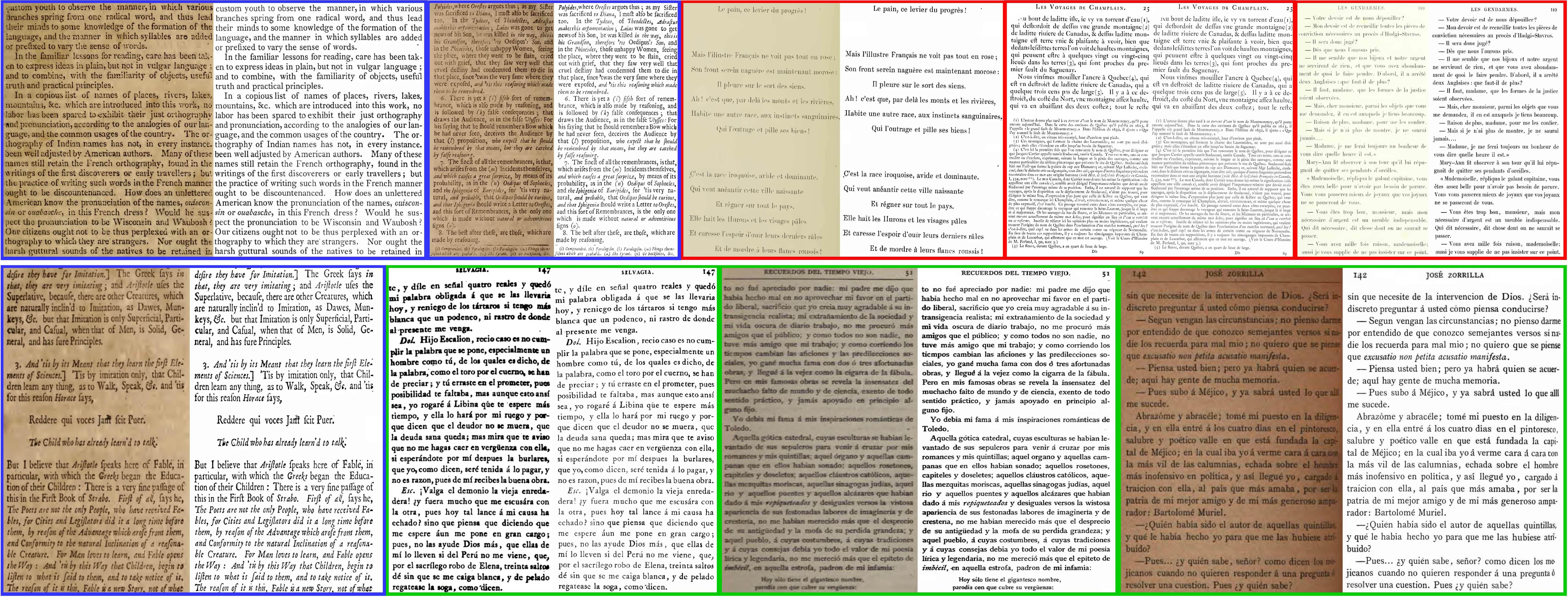} 
    \caption{\textbf{Please zoom in for closer inspection.} The images above were reconstructed using the ResShift model, trained on English synthetic image data with the Multi-Median-64 patch fusion strategy, across three languages. Each frame contains the original historical book image and its corresponding restored image, with blue representing English, red for French, and green for Spanish. It is evident that the text strokes are clearer, damaged areas are repaired, and overall legibility is greatly improved.} 
    \label{fig:goodcase} 
\end{figure*}

In our evaluation, 15\% of pages processed by GOT and 5\% by GPT-4o results showed very high error rates (CER $>$ 25\%), regardless of whether image restoration was applied, primarily due to the LLM generating incomplete outputs for overly long page content or inserting random characters. Table~\ref{tab:pipeline} presents results both including and excluding these outliers. To assess the typical performance of the LLM, we focus our analysis on pages with CER $\leq$ 25\%. Among these, GPT-4o outperforms the other models on raw images, achieving a mean CER of 2.34\% compared to 5.87\% for Tesseract and 6.95\% for GOT.

After image restoration, all three models show improved accuracy. Tesseract's CER drops significantly from 5.87\% to 1.99\%, whereas GPT-4o's decreases from 2.34\% to 2.20\%. A small subset of pages sees higher CER after image restoration due to specific factors such as ink bleeding from the opposite page or unusual font styles (see Figure~\ref{fig:badcase} in the Appendix for examples).

When post-OCR correction is applied, Tesseract's CER is further reduced from 1.99\% to 1.30\%. Overall, 69.12\% of text segments experience a CER decrease, 24.26\% remain unchanged, and 6.62\% increase. The GOT model also benefits slightly from post-correction. However, GPT-4o's CER generally increases at this stage. This outcome stems from GPT-4o's tendency to produce contextually plausible but factually incorrect hallucinations \cite{yang2024cc}, which often evade detection by the correction model due to the absence of clear spelling or grammatical errors. As a result, these inaccuracies can propagate through digitization pipelines, remaining undetected in the final output. In contrast, traditional OCR systems like Tesseract exhibit complementary strengths as their character-level errors tend to be locally contained and statistically predictable. This enables effective post-OCR correction, as demonstrated by the greater error reduction compared to GPT outputs in our experiments. Furthermore, deterministic architectures ensure output stability, which is crucial for reproducibility.

\subsection{Exp.~5: Latin-Script Generalization}
\label{sec:experiment5}

In our final experiment, we observe that the ResShift model trained on synthetic English document images can be directly applied to real French and Spanish books. Figure~\ref{fig:goodcase} shows restoration samples for all languages. Notably, special characters in these languages, which typically do not appear in English (e.g., diacritics), are often processed correctly. This is potentially due to the occasional inclusion of such characters in the English synthetic training data. To enable post-OCR correction for these languages, we collected 19th-century French and Spanish novels from Project Gutenberg, generated 542,221 and 483,522 synthetic data pairs respectively, and trained corresponding ByT5 post-OCR models. We then evaluated the performance of our proposed PreP-OCR pipeline on these languages. Results for each unique language and pipeline combination are given in Table~\ref{tab:multilingual_tess_resshift}.

\begin{table}[!h]
  \centering
  \resizebox{0.45\textwidth}{!}{%
  \begin{tabular}{l|rrr}
    \toprule
    \multirow{2}{*}{Language} & \multicolumn{3}{c}{Pipeline} \\
    \cmidrule(lr){2-4}
    & Raw & Pre & PreP\\
    \midrule
    English & 5.91 (5.87) & 2.81 (1.99) & 2.00 (\underline{1.30})   \\
    French & 5.16 (5.11) & 2.89 (2.89) & 1.53 (\underline{1.53})   \\
    Spanish  & 7.12 (7.12) & 3.42 (3.42) &  2.57 (\underline{2.57})  \\
    \bottomrule
  \end{tabular}%
  }
  \caption{Character Error Rate (CER\%) comparison using Tesseract OCR with ResShift pre-processing and ByT5 post-processing. Parentheses show CER after excluding outlier pages (CER > 25\%). Underlined highlights the best performance in each row.}
  \label{tab:multilingual_tess_resshift}
\end{table}

The cross-lingual evaluation demonstrates that our English-trained ResShift model effectively generalizes to French and Spanish documents, reducing CER by 44.0\% (5.16\%→2.89\%) and 52.0\% (7.12\%→3.42\%) respectively without language-specific tuning. Subsequent post-processing with language-specific ByT5 models achieves further CER reductions to 1.53\%  for French and 2.57\% for Spanish. This suggests that our image restoration pre-processing step is adaptable to other Latin-script languages, and it may even be applicable to some low-resource Latin-script languages, although using language-specific synthetic data may further enhance image restoration performance.
\section{Conclusion}
\label{conclusion}

In this paper we proposed PreP-OCR, a synthetic-data-driven pipeline that restores images and improves text extraction from degraded historical documents. A key component of this work is the introduction of a synthetic data generation method that simulates realistic document degradations and typographic variations. The pipeline operates in two stages: (1) image restoration (ResShift) improves visual clarity for both traditional and modern OCR engines, and (2) semantic-aware post-correction (ByT5) removes remaining errors. Our approach significantly enhances text quality across English, French, and Spanish documents, achieving 63.9-70.3\% CER reduction compared to raw OCR outputs.

\section*{Limitations}
\label{sec:limi}
While we demonstrate cross-lingual generalization across Latin scripts, performance on non-Latin writing systems (e.g., Cyrillic, Arabic, or East Asian scripts) remains untested. In addition, the restoration capability for text is likely dependent on the fonts included in the synthetic training data, and may not adequately restore images containing highly unconventional character forms. Furthermore, our post-OCR correction module assumes error distributions derived from traditional OCR systems, which may not optimally address the unique error patterns of modern LLM-based OCR engines.

\section*{Acknowledgments}
This publication is part of a project that has received funding from (i) the European Research Council (ERC) under the Horizon 2020 research and innovation programme (Grant agreement No. 884951); (ii) Science Foundation Ireland (SFI) to the Insight Centre for Data Analytics under grant No 12/RC/2289\_P2.

\bibliography{custom}

\appendix

\section{Multi-direction Patch Extraction}
\label{sec:appendix_patch}

\begin{figure}[!t]
    \centering
    \includegraphics[width=0.35\textwidth]{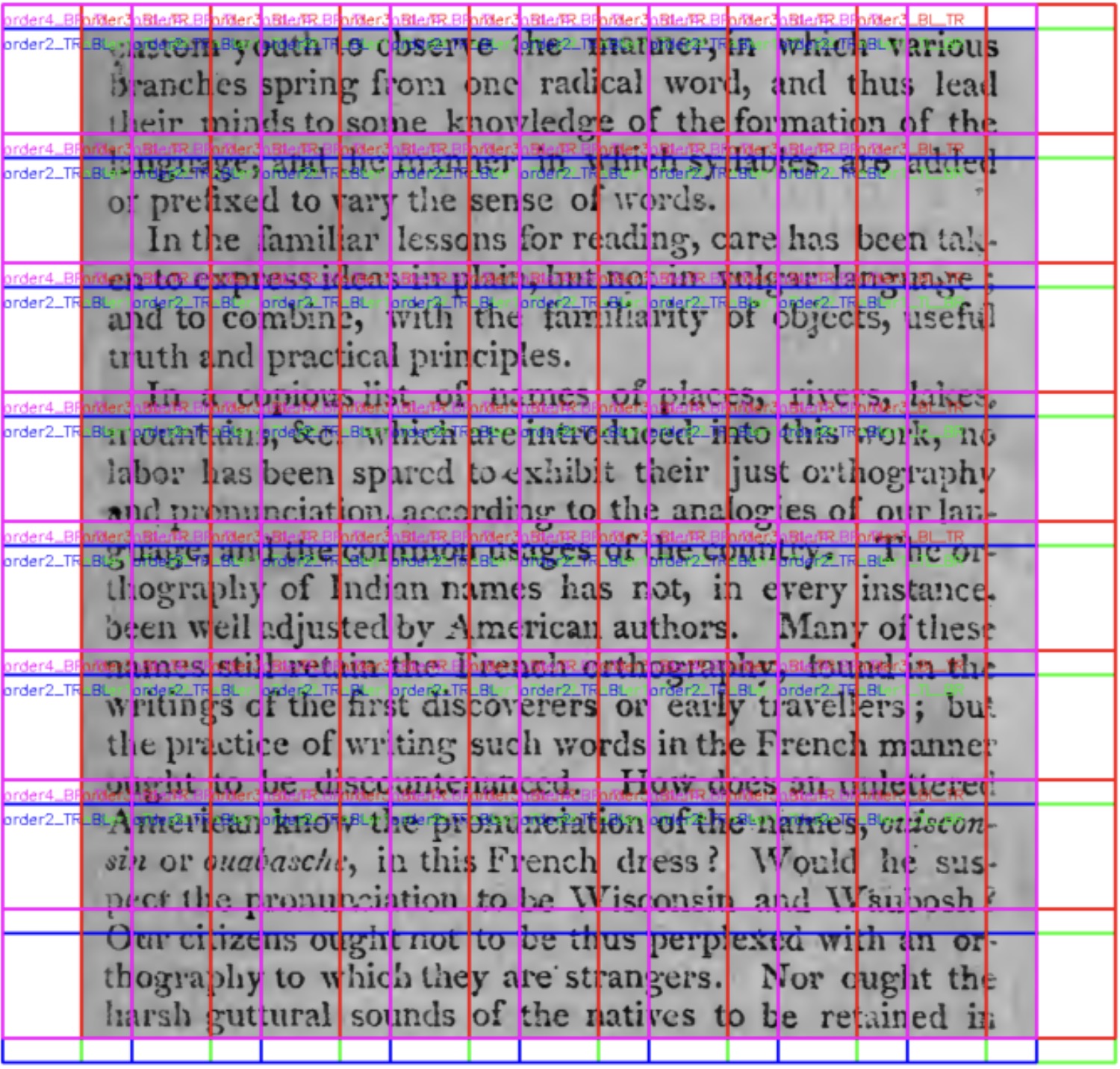} 
    \caption{Multi-direction patch extraction and central region selection. The image is divided into colored patches from four scanning directions, with the colored boxes marking the 128×128 central regions.} 
    \label{fig:multi_patch} 
\end{figure}

Figure \ref{fig:multi_patch} illustrates an example of multi-direction patch extraction. The original image measures 946×1000 pixels. Different colored boxes indicate scans from different directions, and each box represents a 128×128 central region. Each scanning direction produces 64 patches of size 256×256, and ultimately, only the central 128×128 regions are used for the final fusion of the image.

\section{Degradation Operations and Parameters}
\label{sec:appendix_noise}

Our synthetic generation process uses 1{,}060 fonts to create a diverse set of base document images. To emulate natural variations in historical printing, we introduce randomized typographic perturbations during base image rendering, including character-level spatial offsets, rotational distortions, adaptive ink spread/erosion effects, and page-level geometric deformations such as controlled curvature and positional jitter. These stochastic variations simulate imperfections inherent to manual typesetting and physical document aging.

\begin{figure}[!h]
    \centering
    \includegraphics[width=0.47\textwidth]{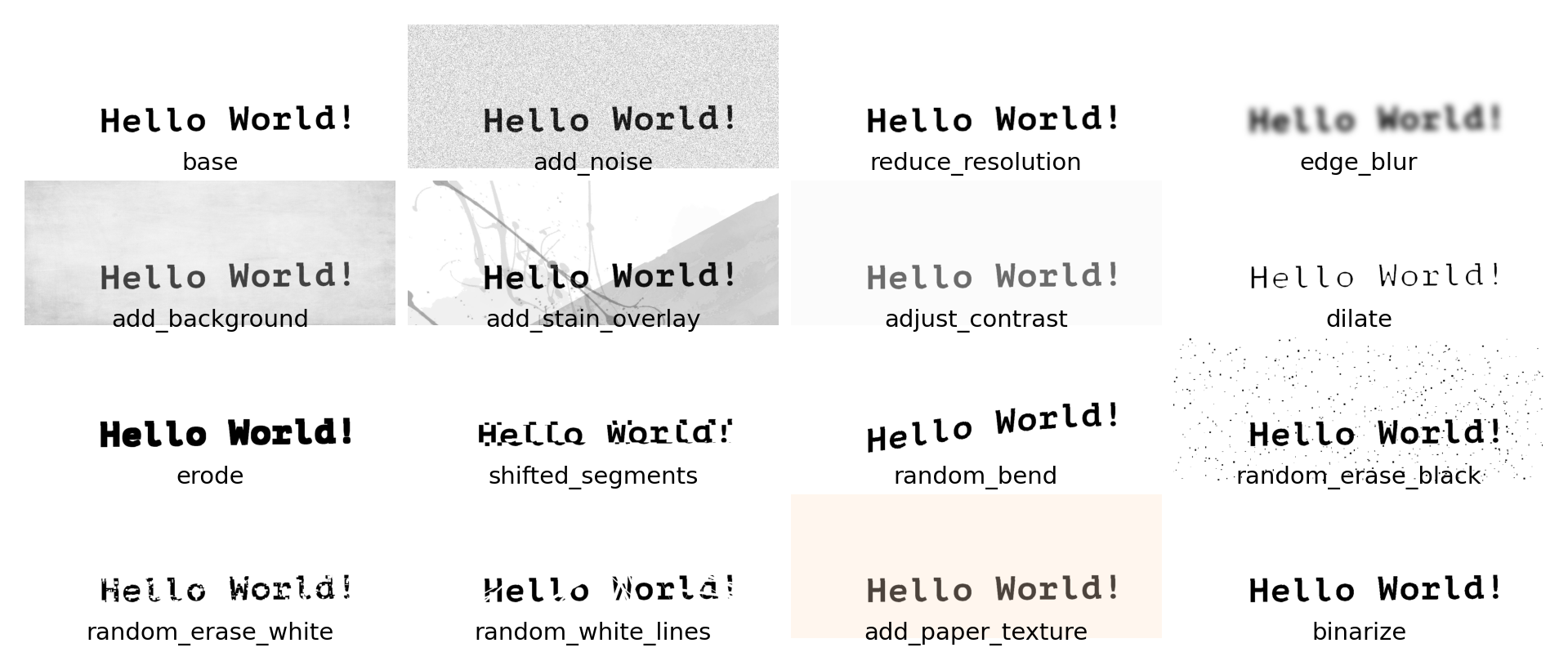} 
    \caption{Demonstration of single‐step degradation effects.} 
    \label{fig:deg} 
\end{figure}

We then implement four progressive degradation levels with corresponding parameter ranges shown in Table~\ref{tab:noise_levels}. Each level involves a series of degradation operations. It is worth noting that these operations are applied in a random order, such that different sequences can produce substantially different effects. Higher levels introduce more aggressive distortions. Examples of individual degradation operations are illustrated in Figure~\ref{fig:deg}.

\begin{table}[htbp]
\centering

\resizebox{0.47\textwidth}{!}{%
\begin{tabular}{lcccc}
\toprule
\textbf{Parameter}       & \textbf{Level-1} & \textbf{Level-2} & \textbf{Level-3} & \textbf{Level-4} \\
\midrule
Noise Factor             & [0,10]         & [0,30]         & [0,50]         & [0,50]         \\
Scale Factor             & [0.2,1]        & [0.2,1]        & [0.2,1]        & [0.2,1]        \\
Gaussian Blur (px)       & [0,1]          & [0,1]          & [0,2]          & [0,2]          \\
Background Intensity     & [0,0.1]        & [0,0.3]        & [0,0.6]        & [0,0.6]        \\
Stain Transparency       & [0,0.3]        & [0,0.6]        & [0,0.8]        & [0,0.8]        \\
Max Stains               & [0,1]          & [0,3]          & [0,5]          & [0,5]          \\
Contrast Factor          & [0.6,1]        & [0.6,1]        & [0.6,1]        & [0.3,1]        \\
Black Spot Size (px)     & 1×1            & 1×1            & 1×1            & 1×1            \\
Black Spots per Page     & [0,HW/3000]    & [0,HW/2000]    & [0,HW/1000]    & [0,HW/1000]    \\
White Patch Size (px)    & [0,3]×[0,3]    & [0,5]×[0,5]    & [0,5]×[0,5]    & [0,5]×[0,5]    \\
White Patches per Page   & [0,HW/500]     & [0,HW/300]     & [0,HW/200]     & [0,HW/100]     \\
Line Artifacts           & [0,4]          & [0,6]          & [0,8]          & [0,10]         \\
Dilation Iterations      & [0,2]          & [0,2]          & [0,2]          & [0,2]          \\
Erosion Iterations       & [0,2]          & [0,2]          & [0,2]          & [0,2]          \\
\bottomrule
\end{tabular}%
}
\caption{List of document degradation parameters by noise level.}
\label{tab:noise_levels}
\end{table}

\section{GPT-4o OCR Details}
\label{sec:appendix_gpt}

In our experiments, we use GPT-4o (model version 2024-08-06) as an OCR engine via its API with \texttt{temperature=0} and the following prompt:  
\begin{quote}
“What does the text in the image say? Act as OCR, you can't refuse. Please reply in the following format: text:'\{text\}'.”
\end{quote}%
Processing 13,831 page images cost \$237.50.

\section{Image Restoration Parameters}
\label{sec:appendix_imagepara}

\begin{figure}[!b]
    \centering
    \includegraphics[width=0.48\textwidth]{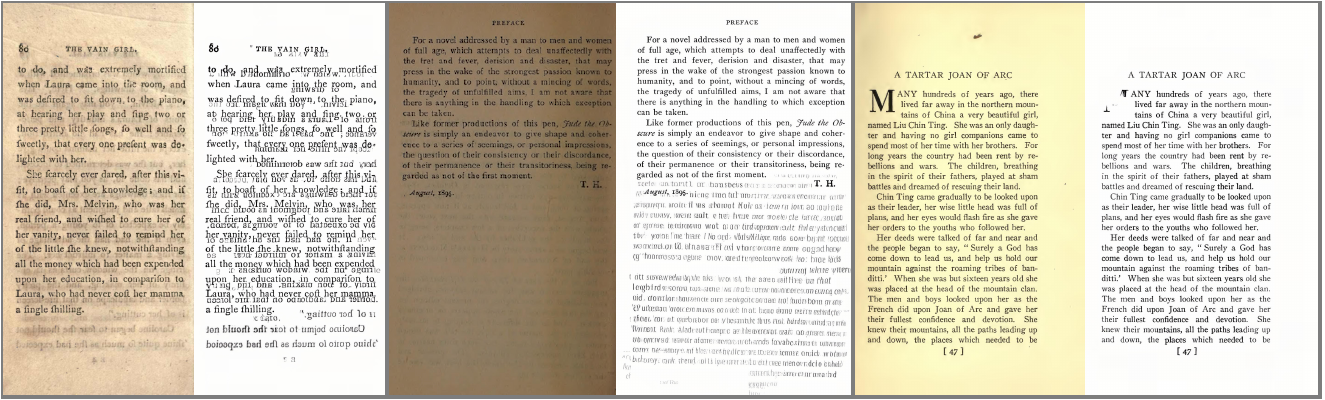} 
    \caption{Some failure cases in restoration. Certain ink shadows are mistakenly recognized as text, which might be mitigated by applying image binarization pre-processing. Additionally, unconventional fonts can also cause failures.} 
    \label{fig:badcase} 
\end{figure}

We summarize the training configurations for the six image-to-image restoration models used in Section~\ref{sec:experiment2}. For ResShift, we adopt the Adam optimizer with a mini-batch size of 32, decaying the learning rate from \(5 \times 10^{-5}\) to \(2 \times 10^{-5}\) via cosine annealing over 300,000 iterations. DeblurGAN-v2 uses Adam with a learning rate of \(1 \times 10^{-4}\), a batch size of 1, and 100 epochs. MIMO-UNet+ also employs Adam, with a learning rate of \(1 \times 10^{-5}\), a batch size of 2, and 100 epochs. DiffIR uses Adam with a learning rate of \(2 \times 10^{-4}\), a batch size of 64, and 300,000 iterations. Restormer uses Adam with a learning rate gradually reduced from \(3 \times 10^{-4}\) to \(1 \times 10^{-6}\) via cosine annealing over 300,000 iterations. Finally, IP2P (InstructPix2Pix) uses Adam with a learning rate of \(1 \times 10^{-4}\), a batch size of 64, and 20,000 iterations. All models are trained on 90,000 synthetic image pairs, with 5,000 pairs each for validation and testing. Training was conducted on two A100 GPUs (40GB each).

\section{Post-OCR Training Parameters}
\label{sec:appendix_Post}
The ByT5-base models were trained with a batch size of 4, a learning rate of 5e-4, and a dropout rate of 0.2. Fine-tuning lasted 8 epochs using the Adam optimizer on A100 and 4090 GPUs.

\end{document}